\documentclass[11pt]{article}

\usepackage[preprint]{acl}  

\usepackage{times}
\usepackage{latexsym}

\usepackage[T1]{fontenc}

\usepackage[utf8]{inputenc}

\usepackage{microtype}
\usepackage{inconsolata}
\usepackage[table,xcdraw]{xcolor}
\usepackage{graphicx}
\usepackage{xspace}
\usepackage{amsmath}
\usepackage{booktabs}
\usepackage{multirow}
\usepackage{makecell}
\usepackage{tcolorbox}
\tcbuselibrary{listings, skins}
\usepackage{soul}
\usepackage{float}
\usepackage{caption}
\usepackage{adjustbox}
\usepackage{bm}
\usepackage{amsmath}
\usepackage{caption}

\usepackage{array}

\title{\tool{}: Iterative Differential Diagnosis with Medical Knowledge Graphs and Information-Guided Inquiring}

\author{
	\textbf{Qipeng Wang\textsuperscript{1}},
	\textbf{Rui Sheng\textsuperscript{2}},
	\textbf{Yafei Li\textsuperscript{2}},
	\textbf{Huamin Qu\textsuperscript{2}},
	\textbf{Yushi Sun\textsuperscript{2}},
	\textbf{Min Zhu\textsuperscript{1}},
	\\
	\textsuperscript{1}Sichuan University, Chengdu, China,
	\textsuperscript{2}HKUST, Hong Kong SAR, China
	\\
	\small{
		\textbf{Correspondence:} \href{mailto:email@domain}{zhumin@scu.edu.cn},
		\href{mailto:ysunbp@connect.ust.hk}{ysunbp@connect.ust.hk}
	}
}

\newcommand{\ie}{i.e.{\xspace}}
\newcommand{\eg}{e.g.,\xspace}

\newcommand{\tool}{MedKGI{\xspace}}
\newcommand{\Doctor}{doctor agent\xspace}
\newcommand{\Patient}{patient agent\xspace}
\newcommand{\Measurement}{measurement agent\xspace}

\newtcblisting{promptlisting}[1][]{%
	enhanced,
	before=\noindent,        
	after=\par,               
	colback=gray!5,
	colframe=gray!40,
	boxrule=0.5pt,
	arc=3pt, 
	fonttitle=\bfseries\sffamily\small,
	coltitle=black,
	colbacktitle=gray!20,
	attach title to upper=\par,
	listing only,
	listing options={
		basicstyle={\ttfamily\small\setlength{\parindent}{0pt}},
		breaklines=true,
		breakatwhitespace=false,
		keepspaces=true,
		columns=fullflexible,
		xleftmargin=0pt,
		xrightmargin=0pt,
		escapeinside={(*@}{@*)},
		resetmargins=true,
		postbreak=\mbox{},
		prebreak={},
		breakindent=0pt,
		breakautoindent=false
	},
	top=2pt, bottom=2pt, left=6pt, right=6pt,
	#1
}

\newcommand{\placeholder}[1]{%
	\colorbox{blue!10}{\texttt{\color{blue!80!black}\{#1\}}}%
}

\begin{document}
\maketitle
\begin{abstract}
Recent advancements in Large Language Models (LLMs) have demonstrated significant promise in clinical diagnosis. 
However, current models struggle to emulate the iterative, diagnostic hypothesis-driven reasoning of real clinical scenarios.
Specifically, current LLMs suffer from three critical limitations: (1) generating hallucinated medical content due to weak grounding in verified knowledge, (2) asking redundant or inefficient questions rather than discriminative ones that hinder diagnostic progress, and (3) losing coherence over multi-turn dialogues, leading to contradictory or inconsistent conclusions.
To address these challenges, we propose \tool{}, a diagnostic framework grounded in clinical practices.
\tool{} integrates a medical knowledge graph (KG) to constrain reasoning to validated medical ontologies, selects questions based on information gain to maximize diagnostic efficiency, and adopts an OSCE-format structured state to maintain consistent evidence tracking across turns.
Experiments on clinical benchmarks show that \tool{} outperforms strong LLM baselines in both diagnostic accuracy and inquiry efficiency, improving dialogue efficiency by 30\% on average while maintaining state-of-the-art accuracy.
\end{abstract}

\section{Introduction} 

Large Language Models (LLMs) are increasingly demonstrating their value as powerful tools for clinical diagnosis~\cite{huatuo2023wang, singhal2025toward, lin2025survey}. However, real-world clinical reasoning is an iterative process in which doctors need to strategically construct diagnostic hypotheses and gather clinical information in a sequential manner to make the final decision. Current LLMs often struggle in such iterative, hypothesis-driven settings due to a fundamental discrepancy between their probabilistic, token-by-token generation and the systematic rigor required for clinical deduction~\cite{eval2024hager}.

\begin{figure}[t]
	\centering
	\includegraphics[width=1\linewidth]{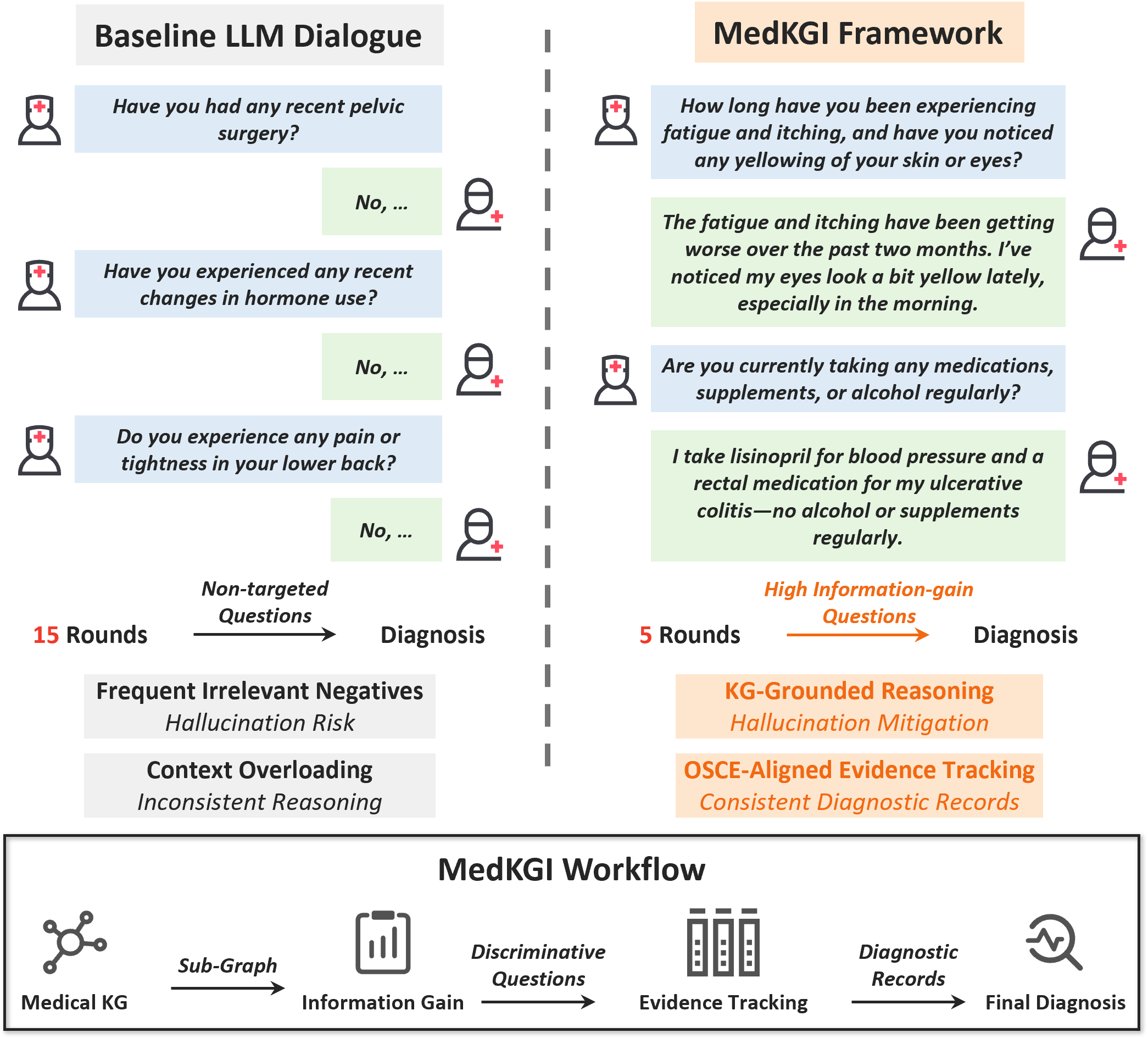}
	\caption{Comparison of diagnostic dialogues and the MedKGI workflow. Left: The dialogues from the baseline LLMs. Right: The dialogues from the proposed \tool{} framework. Bottom: The \tool{} workflow.}
	\label{fig1}
\end{figure}

This gap leads to several critical limitations: a tendency to produce \textit{hallucinations} by prioritizing plausible patterns over verified knowledge~\cite{trust2025zhu}; \textit{ineffective questioning} due to the lack of an explicit reasoning framework~\cite{alfa2025li}; and \textit{context overloading} in multi-turn dialogues caused by their associative nature~\cite{context2024savage}. As illustrated in Figure~\ref{fig1} (left), these limitations often result in redundant and non-strategic diagnostic dialogues by baseline LLMs, in contrast to the structured trajectory of rigorous medical reasoning.

To bridge this gap, we ground our method in established differential diagnosis frameworks rather than language patterns. \textit{Differential diagnosis} is inherently an iterative process of systematically weighing competing hypotheses against clinical evidence \cite{defineddx}, which is the opposite of associative LLM reasoning. Accordingly, we integrate three key principles:

\begin{itemize}
	\item \textbf{Knowledge-Anchored Hypothesis Generation:} Inspired by the clinical practice of grounding initial differentials in established medical knowledge \cite{kg2025zuo}, we integrate a medical knowledge graph (KG) to generate diagnostic candidates based on verified disease–symptom relationships.
	\item \textbf{Strategic Uncertainty Reduction:} Following the differential diagnosis principle of prioritizing high-yield findings, we adopt an information gain-based questioning strategy \cite{uncertainty2025liu} to select the most discriminative questions, thereby minimizing diagnostic uncertainty.
	\item \textbf{Iterative Evidence Refinement:} To simulate a doctor’s belief updating process as new evidence emerges, we implement a state-tracking mechanism that maintains a coherent diagnostic record, enabling consistent hypothesis management while mitigating context overloading in long dialogues \cite{reasoining2024xu}.
\end{itemize}

Building on these principles, we propose \tool{}, a diagnostic reasoning framework designed to emulate the systematic inquiry of human clinicians within the LLM paradigm.
As shown in Figure~\ref{fig1}, \tool{} integrates a medical knowledge graph to anchor all diagnostic reasoning in verified medical ontologies, thereby mitigating hallucinations. Building on this grounded knowledge (sub-graph), it employs an information gain–based question selection strategy. This strategy evaluates candidate questions by their expected reduction in diagnostic uncertainty, enabling \tool{} to prioritize the most discriminative inquiries and optimize diagnostic efficiency. Finally, \tool{} adopts the Objective Structured Clinical Examination (OSCE) format \cite{OCSE2024cushing} to maintain a structured diagnostic state. This state tracks and updates accumulated evidence across dialogue turns, which mitigates context overloading and ensures reasoning consistency.

Our key contributions are:

\begin{itemize}
	\item \textbf{A Systematic, Hypothesis-Driven Diagnostic Framework:} We propose \tool, a novel framework that explicitly models the iterative, hypothesis-driven process of differential diagnosis, bridging the gap between LLMs' generative nature and the analytical rigor of differential diagnosis.
	\item \textbf{Knowledge-Anchored \& Strategically Optimized Reasoning:} \tool{} uniquely integrates a medical knowledge graph to prevent hallucinations and employs an information gain–based questioning strategy to maximize diagnostic efficiency, grounding reasoning in verified ontologies.
	\item \textbf{Superior Empirical Performance:} Extensive experiments show that \tool{} outperforms state-of-the-art baselines, achieving high diagnostic accuracy while improving dialogue efficiency by 30\%.
\end{itemize}

\section{Related Work}

\begin{figure*}[htbp]
	\centering
	\includegraphics[width=1\linewidth]{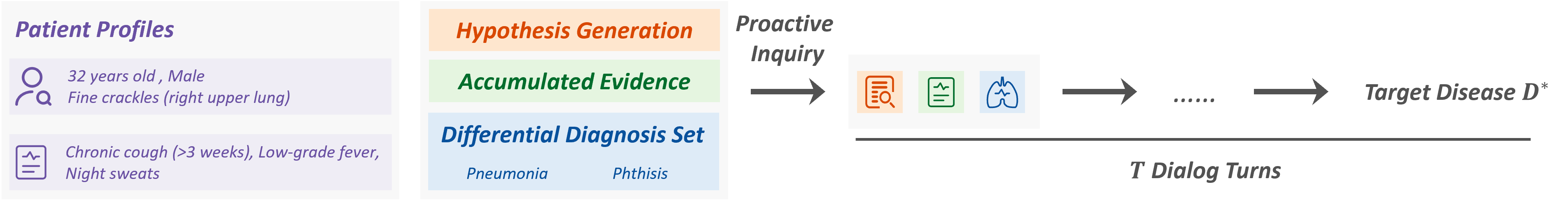}
	\caption{
		An illustration of the iterative hypothesis refinement process and its corresponding clinical differential diagnosis example.
	}
	\label{fig2}
\end{figure*}

Clinical dialogue involves dynamic, multi-turn information exchange and hypothesis refinement~\cite{sequential2025nori}. 
We categorize existing approaches into LLM-driven sequential diagnosis, knowledge-augmented frameworks, and agent-based clinical frameworks.

\textbf{LLM-Driven Sequential Diagnosis.} 
Early methods primarily leverage LLMs' reasoning capabilities, enhanced through fine-tuning or reinforcement learning (RL) to improve medical diagnosis.
Chain-of-Thought (CoT) prompting has been widely adopted to elicit diagnostic reasoning~\cite{cot2025dai}. 
Domain-specialized models for diagnosis like Huatuo~\cite{huatuo2023wang} and Meditron~\cite{meditron2023chen} are pre-trained on medical corpora.
AgentClinic simulates doctor–patient interactions but relies on static prompting without dynamic evidence tracking~\cite{agentclinic2024samuel}.
Recent works have focused on inquiry strategies: MedAgent \cite{medagent2025kim} formulates diagnosis as multi-agent collaboration while PATIENCE \cite{patience2025zhu} incorporates Bayesian active learning for interactive questioning.
However, these model-centric approaches often struggle with hallucinations and struggle with precision in open-ended, multi-turn scenarios due to a lack of external grounding~\cite{kg2025zuo}.

\textbf{Knowledge-Augmented Approaches.} 
To mitigate the limitations of pure LLM-based reasoning, recent work has integrated external knowledge. 
RAG-based methods like MRD-RAG \cite{mrdrag2025sun} leverages the tree-structure medical KG for differential diagnosis, while ClinicalRAG \cite{clinicalrag2024lu} fuses structured and unstructured medical knowledge. 
Beyond retrieval, some methods explicitly model diagnostic reasoning over KGs using search or planning.
For instance, Unit of Thought (UoT)~\cite{uot2024hu} decomposes clinical reasoning into discrete, verifiable knowledge units grounded in a KG.
However, these approaches treat evidence retrieval statically, lacking dynamic state-tracking for handling diagnostic contexts~\cite{adaptive2025wang}.

\textbf{Agent-based Clinical Frameworks.}
Multi-agent systems simulate clinical workflows by decomposing tasks across specialized agents for symptom collection, evidence retrieval, and reasoning.
DDO \cite{ddo2025jia} uses a diagnosis agent, a strategy agent, and a patient agent for stage-specific inquiry, while MeDDxAgent \cite{meddxagent2025rose} integrates a control agent, a history agent, and a knowledge agent to simulate clinical diagnostic processes with external knowledge. 
MEDIQ~\cite{mediq2024li} introduces a query-planning agent that prioritizes questions based on symptom severity, CoD~\cite{cod2025chen} coordinates diagnostic agents through a consensus-driven protocol, and DoctorAgent-RL \cite{doctorrl2025feng} models consultations as an RL process under uncertainty. 
Despite these advances, existing multi-agent frameworks lack criteria for question selection, relying on heuristic role-playing rather than information-theoretic objectives to optimally reduce diagnostic uncertainty~\cite{agent2025chen}.

\textbf{Summary.}
Existing approaches provide flexibility, factual accuracy, and workflow simulation, there is no single existing approach that effectively unifies: (1) KG-grounded reasoning, (2) structured state tracking for context management, and (3) information-theoretic inquiry optimization. 
Our \tool{} framework addresses these challenges by integrating KGs with information gain-driven selection within a structured state tracking mechanism.

\section{Problem Definition}

The multi-step clinical diagnosis can be modeled as an iterative decision-making process that refines hypotheses over $T$ dialogue turns (Figure~\ref{fig2}).
The process begins with the patient profile $\mathcal{P}$, which includes demographics and chief complaints. 
Over a sequence of dialogue turns $t$, the framework maintains a dynamic clinical state. 
At turn $t$, given $\mathcal{P}$ and accumulated evidence $\mathcal{E}_t$, the objective is to estimate the posterior probability for each candidate disease $D \in \mathcal{D}_t$, where $\mathcal{D}_t = \{D_1, D_2, \dots , D_n \}$ through evidence collection.
The proactive symptom inquiry is defined as identifying the optimal inquiry $s$ that maximizes the expected reduction in diagnostic uncertainty. This mechanism enables the framework to iteratively generate and refine $D_t$ until the target disease $D^*$ is reached:

$$P(D \mid \mathcal{P}, \mathcal{E}_t) \to \delta(D, D^*)$$

\section{Methodology}

\begin{figure*}[t]
	\centering
	\includegraphics[width=1\linewidth]{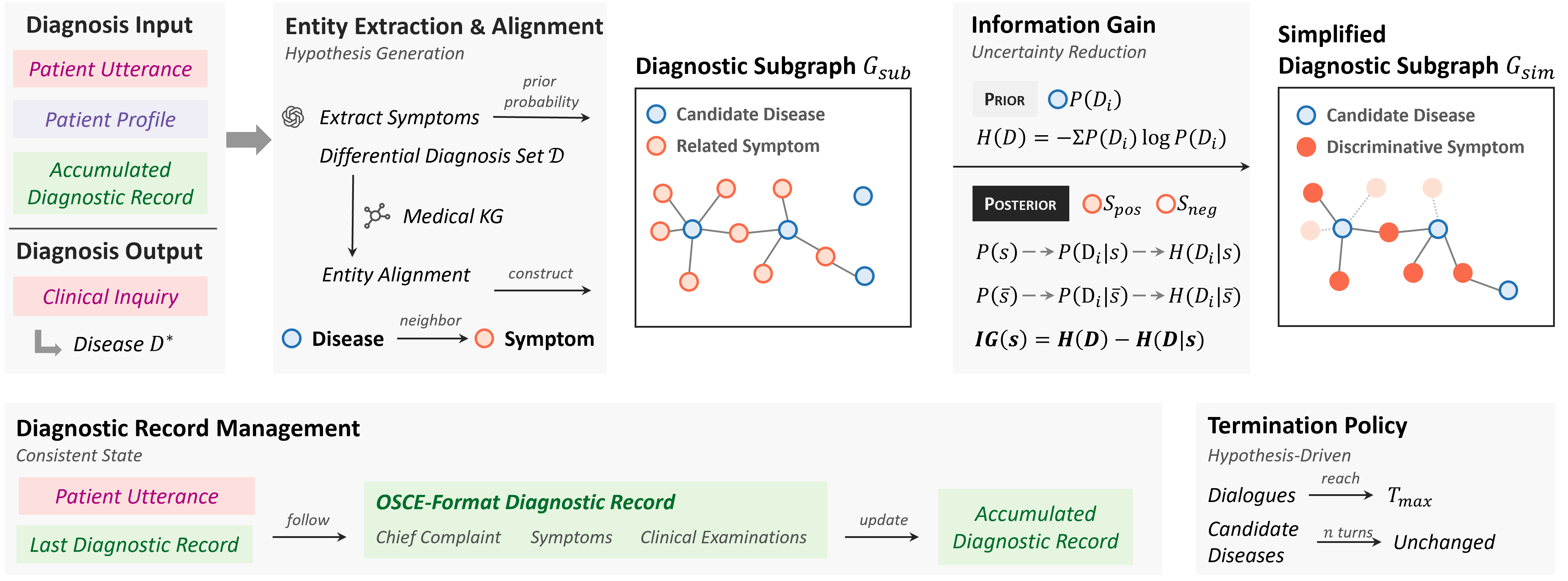}
	\caption{
		An overview of \tool{} framework. Given a patient's chief complaint, \tool{} iteratively refines differential diagnosis through (1) medical knowledge graph alignment, (2) information gain–driven symptom inquiry to minimize diagnosis uncertainty, and (3) OSCE-aligned diagnostic records for coherent evidence tracking. A hypothesis-driven termination policy ensures diagnostic efficiency.
	}
	\label{fig3}
\end{figure*}

As illustrated in the differential diagnosis scenario (Figure~\ref{fig2}), given patient profiles and symptoms, the objective is to narrow a differential diagnosis set toward the target disease $D^*$ through sequential evidence collection.
Unlike static classification, this scenario requires actively navigating a hypothesis space, simulating a doctor's cognitive process~\cite{bayesian2024polotskaya}.
Specifically, we formulate the iterative refinement process (Figure~\ref{fig3}) as a knowledge-guided active diagnostic framework, where a \Doctor iteratively refines the differential diagnosis set by strategically gathering discriminative evidence.

To realize this iterative and knowledge-driven process, we design the \tool{} workflow integrating three components: 
(1) \textbf{Entity Extraction \& Alignment:} maps the diagnosis input and generated hypothesis to a medical KG, constructing a diagnostic subgraph grounded in clinically validated disease-symptom relationships to mitigate hallucinations,
(2) \textbf{Information Gain-Based Inquiry:} calculates information gain to identify discriminative symptoms, ensuring each clinical inquiry maximally reduces diagnostic uncertainty and improves diagnostic efficiency,
(3) \textbf{OSCE-Aligned Diagnostic Record Management:} organizes accumulated evidence into an OSCE-format diagnostic record, maintaining a consistent state to prevent context overloading.

\subsection{Diagnosis Workflow of \tool{}}

At any diagnostic turn $t$, the \Doctor receives the following as input:

\begin{itemize}
	\item The \textit{Patient Profile} (\eg age, sex, and chief complaints) provided at the beginning;
	\item The \textit{Patient's Recent Utterance}, which may contain new symptoms or responses to prior questions;
	\item The \textit{Accumulated Diagnostic Record}, a structured OSCE-aligned summary (detailed in Section \ref{record}) containing confirmed symptoms, medical history, and examination findings up to turn $t-1$.
	
\end{itemize}

Based on this context, the \Doctor generates two key outputs:

\begin{itemize}
	\item A natural language \textit{Clinical Inquiry} to elicit discriminative evidence;
	\item A \textit{Differential Diagnosis Set} outputted by \Doctor;
\end{itemize}

To ensure efficiency and diagnostic precision,  diagnostic process concludes when one of the following termination conditions is met 

\begin{itemize}
	\item \textbf{Turn Limit.} If the dialogue reaches $T_{max}$ turns, the \Doctor must issue a final diagnosis.
	\item \textbf{Stagnation Detection.} If the differential diagnosis set remains unchanged for $n$ consecutive turns, \Doctor is prompted to seek evidence that could refute the current hypothesis. If no such symptom exists, the \Doctor outputs the final diagnosis.
\end{itemize}

\subsection{Entity Extraction and Knowledge Graph Alignment}
At each turn, the LLM first proposes a preliminary differential diagnosis set, which is then mapped to the medical KG.
To align medical terms mentioned in the patient utterance with standardized entities in the KG, we implement a multi-stage alignment pipeline:

\begin{itemize}
	\item \textbf{Exact Matching.} For standard medical terms in the dialogue, we query the KG for an entity that exactly matches the candidate disease name.
	\item \textbf{Edit-Distance Matching.} For minor spelling variations or errors, we apply the Levenshtein Distance \cite{Levenshtein1965BinaryCC} to identify approximate matches, allowing a maximum edit distance of 3.
	\item \textbf{Semantic Embedding Matching.} For conceptually equivalent but lexically divergent expressions, we leverage the pre-trained PubMedBERT \cite{pubmedbert2021} to generate vector embeddings for the candidate disease name and all KG disease entity names. We calculate cosine similarity and retrieve the top-ranked entity, discarding matches with a similarity score below a threshold $\tau=0.85$ to ensure alignment quality.
\end{itemize}

\subsection{Information Gain-Based Symptom Selection}

Once the candidate diseases are mapped to the KG, we construct a task-specific diagnostic subgraph $G_{sub} = (V_{sub}, E_{sub})$, comprising the current differential diagnosis set and their directly connected symptom nodes.
To strategically reduce diagnostic uncertainty, \tool{} selects symptom queries that maximize information gain \cite{informationgain1986}. This selection is made over the diagnostic subgraph $G_{sub}$ and is conditioned on the patient's reported positive and negative symptoms ($S_{pos} \text{, } S_{neg}$).

\subsubsection{Posterior Disease Probability Estimation}

First, we establish the context by constructing a diagnostic subgraph $G_{sub}=(V_{sub}, E_{sub})$, where $V_{sub}={\textstyle \bigcup_{i=1}^{n}} \{D_i\} \cup N(D_i)$ consists of the differential diagnosis set $\mathcal{D} = \{ D_1, D_2, \dots, D_n \} $ and all symptoms connected to them in the KG: $N(D_i)=\{ s \in V_{KG} \vee s \in \mathcal{S} : (D_i, s) \in E_{KG} \}$.
We initialize the prior probability of each candidate disease $D_i$ based on its average semantic similarity to the confirmed symptoms $S_{pos}$ extracted from the current dialogue:
$$P(D_i)=\frac{1}{\mid S_{pos} \mid} \sum_{S \in S_{pos}} semantic\_sim(s, D_i) $$
where $semantic\_sim(\cdot, \cdot)$ denotes cosine similarity between the PubMedBERT \cite{pubmedbert2021} embeddings of symptom $s \in S_{pos}$ and disease $D_i$.

Given the accumulated observed symptoms $S_{pos}$ and $S_{neg}$, we update disease beliefs over candidate disease $\mathcal{D}$ using Bayes' theorem:
$$P(D_i \mid S_{pos}, S_{neg}) = \frac{P(S_{pos}, S_{neg} \mid D_i) \cdot P(D_i)}{P(S_{pos}, S_{neg})}$$

Assuming conditional independence among symptoms given a disease, the likelihood factorizes as:
$$P(S_{pos}, S_{neg})=\sum_{j=1}^{n} P(S_{pos}, S_{neg} \mid D_j) \cdot P(D_j)$$ 
where we adopt a uniform conditional probability model: $P(s, \mid D_i)=1 / \left | N(D_i) \right |$ for all symptoms $s \in N(D_i)$.

\subsubsection{Information Gain Computation}

For the current differential disease set $\mathcal{D}$ with posterior probabilities $P(D_i \mid S_{pos}, S_{neg})$, we compute the prior diagnostic uncertainty using the Shannon Entropy:  
$$ H(\mathcal{D})=-\sum^{n}_{i=1}P(D_i)\log{P(D_i)} $$
For any symptom $s$, we compute its marginal probability and the resulting posterior disease distributions: 
$$P(s)=\sum^{n}_{i=1}P(s \mid D_i)P(D_i) \text{,} $$
$$P(D_i \mid s)=\frac{P(s \mid D_i)P(D_i)}{P(s)}$$

Finally, the Information Gain of asking about symptom $s$ is defined as the expected reduction in entropy: 
$$ IG(s)=H(\mathcal{D})-H(\mathcal{D} \mid s) $$ 
where $H(\mathcal{D} \mid s)$ represents the expected conditional entropy after observing symptom $s$:
$$ H(\mathcal{D} \mid s) = P(s)H(\mathcal{D} \mid s^+) + P(\neg s)H(\mathcal{D} \mid s^-) $$ 
and $H(\mathcal{D} \mid s^+)$ and $H(\mathcal{D} \mid s^-)$ are the entropies if the symptom is observed positive or negative, respectively.
The \Doctor then selects the top-$k$ symptoms with the highest $IG(s)$ for strategic inquiry, ensuring that each subsequent question maximally reduces uncertainty.
Compared to methods that rely on predefined question templates or fixed inquiry sequences, \tool{} dynamically adapts questions based on the evolving diagnostic hypothesis, enabling more targeted and efficient information gathering.

\subsection{Consistent State by Diagnostic Record Management}
\label{record}

To support coherent and consistent reasoning, we employ the LLM to generate and maintain a structured diagnostic record in JSON format, aligned with the Objective Structured Clinical Examination (OSCE) standard. 
At the beginning of each dialogue session, we initialize an empty diagnostic record following a predefined schema, including chief complaint, symptoms, and recent medical examinations.

At each turn, \tool{} takes the latest diagnostic record and patient utterance as input.
Then, \tool{} outputs an updated diagnostic record that integrates new evidence while preserving clinical context. This accumulated diagnostic record prevents context overloading across turns, which commonly occurs when context windows accumulate redundant information in vanilla prompting methods \cite{agentclinic2024samuel}.

\section{Evaluation}

\subsection{Experiment Setup}

\textbf{Datasets.}
We conducted experiments on two medical QA benchmarks: MedQA \cite{medqa2021jin} and CMB \cite{cmb2024wang, cmedbenchmark}.
To further assess multi-modal clinical reasoning, we additionally introduce a dataset of real-world cases from the NEJM Image Challenge\footnote{https://www.nejm.org/image-challenge}.
We followed the settings of AgentClinic~\cite{agentclinic2024samuel} to simulate the multi-agent medical consultation scenarios based on the cases in MedQA and CMB. We denote the processed datasets as agent-MedQA and agent-CMB.

\begin{table*}[t]
	\caption{Comprehensive evaluation of \tool{} across three benchmarks agent-MedQA, agent-CMB, and NEJM using Qwen3-8B, Llama3.1-8B-Instruct, Qwen3-VL-8B-Instruct as base language models. 
		All methods are evaluated with a maximin dialogue round of 20.
		Reported metrics including average dialogue rounds (\textit{Rounds} \boldsymbol{$\downarrow$}) and diagnostic accuracy (\textit{Acc} (\%) \boldsymbol{$\uparrow$}).
		Methods marked with \textsuperscript{*} employ specialized LLMs (\ie HuatuoGPT-o1, Meditron-7B, and DiagnosisGPT-7B) rather than the base LLM used in our unified evaluation. 
		Best results per column are \textbf{bolded}; second-best are \underline{underlined}.}
	\label{table1}
	\setlength{\tabcolsep}{3pt}
	\renewcommand{\arraystretch}{1.4}
	\small
	\centering
	\begin{tabular}{lcccccccccc}
		\toprule 
		\multicolumn{1}{l|}{} & \multicolumn{4}{c|}{\textbf{agent-MedQA}}                                                 & \multicolumn{4}{c|}{\textbf{agent-CMB}}                                                  & \multicolumn{2}{c}{\textbf{NEJM}}                 \\ \cmidrule(l){2-11} 
		\multicolumn{1}{l|}{\multirow{1}{*}{\textbf{Baselines}}}                           & \multicolumn{2}{c}{Qwen3-8B}   & \multicolumn{2}{c|}{Llama3.1-8B-Instruct} & \multicolumn{2}{c}{Qwen3-8B}   & \multicolumn{2}{c|}{Llama3.1-8B-Instuct} & \multicolumn{2}{c}{Qwen3-VL-8B-Instruct} \\ \cmidrule(l){2-11} 
		\multicolumn{1}{l|}{}                           & \textit{Rounds}        & \textit{Acc (\%)}        & \textit{Rounds}            & \multicolumn{1}{c|}{\textit{Acc (\%)}}             & \textit{Rounds}        & \textit{Acc (\%)}       & \textit{Rounds}             & \textit{Acc (\%)}            & \textit{Rounds}             & \textit{Acc (\%)}             \\ \midrule
		\multicolumn{11}{l}{\cellcolor[HTML]{F2F2F2}\textbf{LLM-Based}}                                                                                                                                                                                               \\
		\multicolumn{1}{l|}{AgentClinic}                & 11.32         & 59.43          & 10.37                  & \multicolumn{1}{c|}{50.00}                     & 10.00         & 58.28          &  11.23                  & \multicolumn{1}{c|}{59.60}                   & 13.28                   & 54.55                    \\
		\multicolumn{1}{l|}{CoT}                        & 18.92         & 24.52          & 17.46                  & \multicolumn{1}{c|}{34.90}                     & 18.32         & 43.05          & 16.33                   & \multicolumn{1}{c|}{50.33}                   & 16.46                   & 50.91                    \\
		\multicolumn{1}{l|}{\begin{tabular}[c]{@{}l@{}}Huatuo\textsuperscript{*}\\ [-5pt] \textit{\scriptsize(HuatuoGPT-o1)}\end{tabular}}                    & 16.70         & 56.60          & -                 & \multicolumn{1}{c|}{-}                    & 16.56         & 62.25          & -                  & \multicolumn{1}{c|}{-}                  & -                  & -                   \\
		\multicolumn{1}{l|}{\begin{tabular}[c]{@{}l@{}}Medical-CoT\textsuperscript{*}\\ [-5pt] \textit{\scriptsize(MediTron-7B)}\end{tabular}}               & 18.94         & 60.37          & -                 & \multicolumn{1}{c|}{-}                    & 18.34         & \underline{66.23}          & -                  & \multicolumn{1}{c|}{-}                  & -                  & -                   \\ \midrule
		\multicolumn{11}{l}{\cellcolor[HTML]{F2F2F2}\textbf{KG-Based}}                                                                                                                                                                                                \\
		\multicolumn{1}{l|}{MCTS-BT}                    & 14.20         & 45.28          & 14.98                  & \multicolumn{1}{c|}{39.62}                    & 13.78         & 54.97          & 14.21                    & \multicolumn{1}{c|}{42.38}                   & \underline{11.50}                   & 56.36                    \\
		\multicolumn{1}{l|}{MCTS-MV}                    & 14.63         & 53.77          & 12.86                  & \multicolumn{1}{c|}{\underline{52.83}}                     & 14.77         & 60.26          & 12.91                   & \multicolumn{1}{c|}{56.95}                   & 13.95                   & 67.27                    \\
		\multicolumn{1}{l|}{UoT}                        & 11.47         & 54.71          & 11.01                  & \multicolumn{1}{c|}{51.89}                     & 10.71         & 64.28          & 10.96                   & \multicolumn{1}{c|}{58.94}                   & 12.32                   & 54.55                    \\ \midrule
		\multicolumn{11}{l}{\cellcolor[HTML]{F2F2F2}\textbf{Agent-Based}}                                                                                                                                                                                             \\
		\multicolumn{1}{l|}{MediQ}                      & 13.80          & \underline{61.32}          & 13.85                & \multicolumn{1}{c|}{49.06}                     & 13.76         & 65.56          & 15.01                   & \multicolumn{1}{c|}{\textbf{60.93}}                   & 14.48                   & 65.45                    \\
		
		\multicolumn{1}{l|}{DDO}                        & 17.27         & \underline{61.32}          & 18.39                  & \multicolumn{1}{c|}{50.94}                     & 17.38         & 63.58          & 18.01                   & \multicolumn{1}{c|}{\textbf{60.93}}                   & 17.91                   & \textbf{70.73}                    \\
		\multicolumn{1}{l|}{MEDDxAgent}                 & 16.44         & 60.38          & 16.02                  & \multicolumn{1}{c|}{49.06}                     & 16.09         & 61.59          & 16.48                   & \multicolumn{1}{c|}{57.62}                   & 16.36                   & 65.45                    \\
		
		\multicolumn{1}{l|}{\begin{tabular}[c]{@{}l@{}}CoD\textsuperscript{*}\\ [-5pt] \textit{\scriptsize(DiagnosisGPT-7B)}\end{tabular}}                        & 13.32         & 56.60          & -                  & \multicolumn{1}{c|}{-}                     & 11.99         & 28.50          &  -                  &  \multicolumn{1}{c|}{-}                  & -                   & -                    \\\midrule
		\multicolumn{11}{l}{\cellcolor[HTML]{F2F2F2}\textbf{SFT-Based}}                                                                                                                                                                                               \\
		\multicolumn{1}{l|}{SFT}                       & 11.51         & 51.89          & 11.21                 & \multicolumn{1}{c|}{49.06}                    & 12.04         & 59.60          & \underline{9.95}                  & \multicolumn{1}{c|}{52.98}                  & -                  & -                   \\
		\multicolumn{1}{l|}{SFT-GT}                     & \underline{9.35}          & 50.94          & \underline{10.27}                  & \multicolumn{1}{c|}{40.57}                     & \underline{9.93}          & 55.63          & 10.26                   & \multicolumn{1}{c|}{51.66}                   & -                  & -                   \\ \midrule
		\rowcolor[HTML]{ffefe0}
		\multicolumn{1}{l|}{\textbf{Ours}}              & \textbf{9.11} & \textbf{69.81} & \textbf{10.20}                  & \multicolumn{1}{c|}{\textbf{53.77}}            & \textbf{9.13} & \textbf{68.21} & \textbf{9.72}                   & \multicolumn{1}{c|}{\underline{60.26}}                   & \textbf{10.53}                   & \underline{69.09}                    \\ \bottomrule
	\end{tabular}
\end{table*}

\textbf{Baselines.} 
We compared \tool{} against 12 baselines across four categories: (1) Dialog-Based Methods: AgentClinic \cite{agentclinic2024samuel}, CoT (Chain-of-Thought), Huatuo \cite{huatuo2023wang} and Meditron \cite{meditron2023chen}; (2) KG-Based Methods: MCTS-BT~\cite{promed2025ding}, MCTS-MV~\cite{promed2025ding}, UoT \cite{uot2024hu}; (3) Agent-Based Methods: MEDIQ \cite{mediq2024li}, CoD \cite{cod2025chen}, DDO \cite{ddo2025jia}, and MEDDxAgent \cite{meddxagent2025rose}; and (4) SFT-Based Methods including models fine-tuned on domain-specific dialogues. 
Specialized medical LLMs (\eg Huatuo and Meditron) are not evaluated on the NEJM benchmark if they lack the ability of multi-modal analysis. 
Furthermore, SFT and SFT-GT are excluded from NEJM evaluation due to the lack of multimodal dialogue training data required for effective fine-tuning.
A complete description of all individual models and their configurations is provided in Appendix \ref{appendix1}.

\textbf{Implementation.} 
Details of our knowledge graph integration are provided in the Appendix \ref{kg}.
To simulate realistic clinical interactions, we implemented a multi-agent framework with three specialized agents, adapted from AgentClinic~\cite{agentclinic2024samuel}: 

\begin{itemize}
	\item Doctor agent asks up to $T_{max} = 20$ questions.
	\item Patient agent responds only with symptom descriptions and never reveals diagnosis.
	\item Measurement agent simulates the outcome of laboratory tests or medical examinations.
\end{itemize}

We modified inquiry termination criteria and evidence-collection protocols to better align with clinical workflows. Detailed descriptions of the specific prompt engineering for each agent are provided in Appendix \ref{appendix6}.

Our experiment employed Qwen3-8B \cite{qwen32025} and Llama3.1-8B-Instruct \cite{Llama3} for agent-MedQA and agent-CMB, and Qwen3-VL-8B-Instruct \cite{qwen32025} for NEJM. For specialized models (\eg Huatuo and Meditron), we used their architectures.

\textbf{Metrics.} 
Diagnostic accuracy (\textit{acc}): The diagnostic accuracy is quantified by the exact match between the final diagnostic output and the ground truth $D^*$. Higher values indicate a more robust alignment with clinical benchmarks.
Dialogue rounds (\textit{Rounds}): We also record the average dialogue turns to reach a diagnosis for each method. Fewer \textit{Rounds} indicate more efficient diagnosis.

\subsection{Main Result}

Table \ref{table1} presents a comprehensive comparison of \tool{} against state-of-the-art baselines across three medical consultation benchmarks, agent-MedQA, agent-CMB, and NEJM, using multiple backbone LLMs. Our method demonstrates superior performance in both diagnostic accuracy and efficiency.

\textbf{Overall Performance.}
\tool{} achieves superior accuracy across three benchmarks using comparable base models: 69.81\% on agent-MedQA (Qwen3-8B), 68.21\% on agent-CMB (Qwen3-8B), and 69.09\% on NEJM (Qwen3-VL-8B-Instruct). Notably, these results are obtained with the fewest dialogue rounds, 9.11, 9.13, and 10.53 out of a maximum of 20 rounds respectively.

\textbf{Comparison with LLM-based Methods.}
Compared to general LLM-based approaches, \tool{} outperforms methods like AgentClinic and CoT across all benchmarks. It also surpasses specialized medical LLMs (marked with *). For instance, on agent-CMB, \tool{} using Qwen3-8B achieves higher accuracy (68.21\%) than Medical-CoT with MediTron-7B (66.23\%), while doing so in significantly fewer dialogue rounds.

\textbf{Comparison with KG-based and Agent-based Methods.}
Among KG-based methods, \tool{} surpasses even competitive approaches like MCTS-MV and UoT. For instance, on agent-CMB with Qwen3-8B, it achieves 68.21\% accuracy, exceeding UoT’s 64.28\%. In contrast to agent-based approaches such as MediQ, DDO, and MEDDxAgent, \tool{} also demonstrates superior performance. Furthermore, compared to the state-of-the-art method, \tool{} achieves comparable or better accuracy while typically requiring far fewer rounds across all three datasets.

\textbf{Comparison with SFT-based Methods.}
While SFT-based methods achieve competitive dialogue efficiency, their accuracy lags behind that of our method. For example, on agent-MedQA using Qwen3-8B, SFT-GT achieves comparable efficiency (9.35 average rounds vs. our 9.11) but its accuracy (50.94\%) is significantly lower than ours (69.81\%).

\subsection{Analysis of Superior Performance}

The performance of \tool{} generalizes across different backbone LLMs. For example, with Llama3.1-8B-Instruct, our method achieves the highest accuracy on agent-MedQA (53.77\%) and the second-highest on agent-CMB (60.26\%), while consistently requiring the fewest dialogue rounds.
The superiority of \tool{} across diverse benchmarks and LLMs stems from three key factors: knowledge grounding, context-aware reasoning, and efficient inquiry.
First, unlike methods that rely solely on pre-trained LLM knowledge (e.g., AgentClinic, CoT), which may lack structured clinical reasoning, \tool{} integrates a medical knowledge graph (KG). This external grounding enables precise inference and provides a structured hypothesis space for active querying~\cite{medikal2025jia}. Our ablation study (Table \ref{table2}) confirms that removing the KG leads to a significant drop in accuracy (-25.47\% on agent-MedQA).
Second, compared to other KG-based methods that often rely on heuristic metrics for symptom selection, \tool{} selects questions based on information gain that accounts for patient-specific context. This approach avoids both random noise and popularity bias, leading to more discriminative queries~\cite{llmreasoning2025kim}.
Third, in contrast to agent-based methods (e.g., DDO, MEDDxAgent), our framework minimizes redundant interactions by dynamically pruning the candidate symptom set based on information gain and maintaining diagnostic records. This enables \tool{} to achieve diagnosis in fewer rounds while maintaining high symptom coverage.

\begin{table}[t]
	\centering
	\small
	\captionof{table}{Ablation experiments on agent-MedQA using Qwen3-8B, demonstrating the contribution of each component to diagnostic performance.}
	\label{table2}
	\setlength{\tabcolsep}{3pt}
	\renewcommand{\arraystretch}{1.4}
	\begin{tabular*}{\linewidth}{@{\extracolsep{\fill}}lcc@{}}
		\toprule
		\textbf{Method} & \textbf{Rounds} & \textbf{Acc (\%)} \\ \midrule
		{\textit{w/o}} Knowledge Graph & 13.44 & 44.34 \\
		\textit{w/o} Clinical Record & 12.09 & 57.55 \\
		Random node selection & 19.25 & 31.13 \\
		Degree-based node selection & 17.47 & 47.17 \\ \midrule
		\textbf{Ours} & \textbf{9.11} & \textbf{69.81}\\ 
		\bottomrule
	\end{tabular*}
\end{table}

\subsection{Ablation Experiments}
We tested three variants for the ablation study: (1) removing KG integration; (2) disabling the Clinical Record module; and (3) replacing the information gain-based symptom pruning strategy with random or frequency-based alternatives.
As shown in Table \ref{table2}, the full framework consistently achieves the highest diagnostic accuracy. Removing KG integration leads to a significant performance drop, underscoring the critical role of structured external knowledge. Meanwhile, omitting dialogue history summarization results in incomplete patient records, which impairs contextual coherence over multi-turn interactions. Finally, both random and frequency-based symptom pruning strategies result in lower accuracy than our information-gain approach, confirming that targeted, discriminative symptom selection is essential.
Together, these findings validate the necessity of each key component in our design.

\subsection{Hyper parameter Selection}
\begin{figure}[t]
	\centering
	\includegraphics[width=\linewidth]{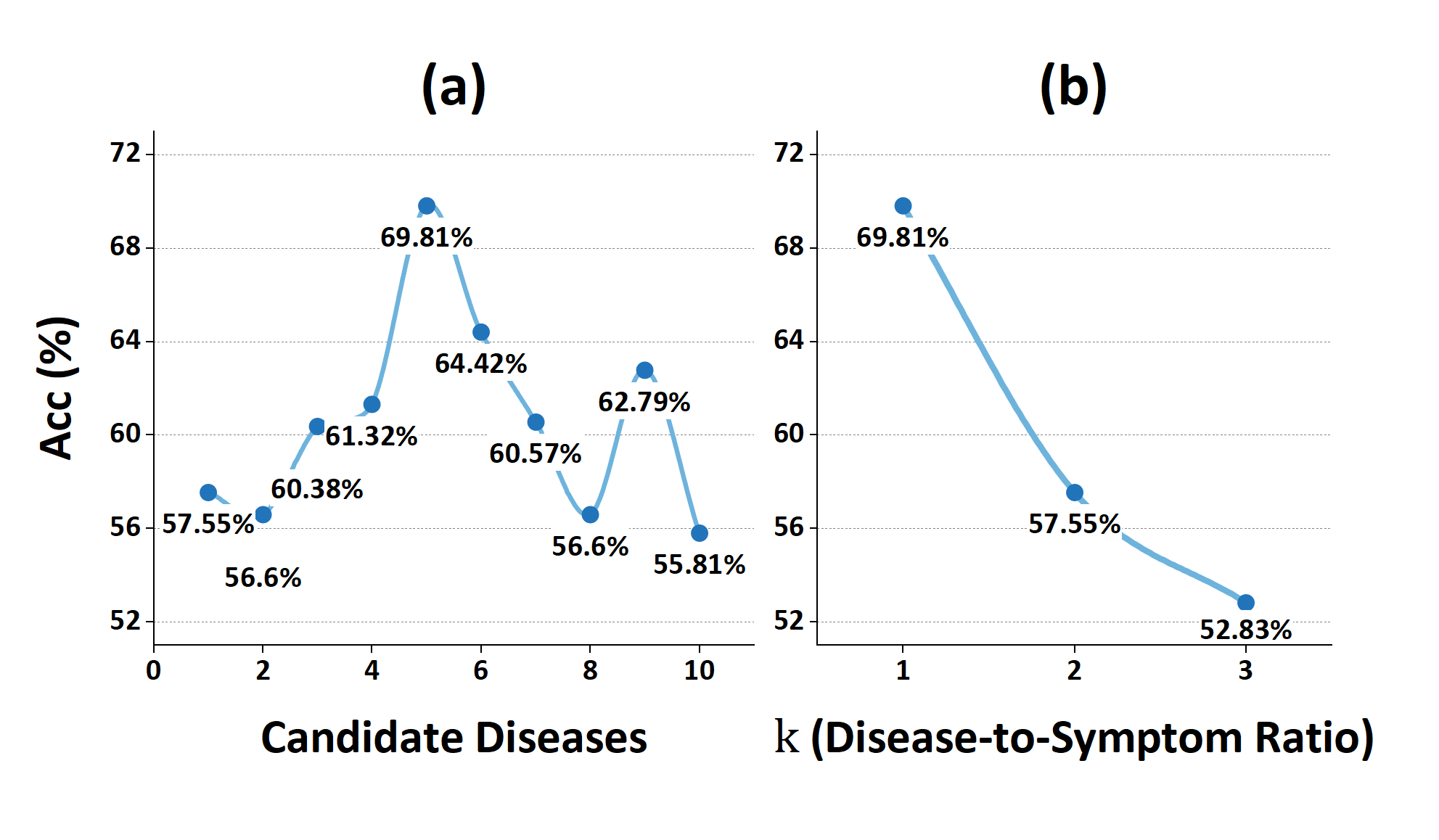}
	\captionof{figure}{(a) Impact of candidate disease settings on accuracy. (b) Effect of $k$ on Accuracy where $k$ is defined as the ratio of the number of candidate diseases to the number of related symptoms from the KG.}
	\label{fig4}
\end{figure}

We performed controlled experiments to determine the optimal settings for two key hyperparameters.
First, we examined how the number of candidate diseases affects accuracy. Figure \ref{fig4}(a) shows that accuracy peaks at 5 candidates and declines with more, as low-relevance candidates introduce noise. In practice, we recommend finding the optimal value by testing on a small sampled dataset.
Second, we tested symptom sampling by defining $k$ as the average symptoms per candidate disease. Figure \ref{fig4}(b) indicates optimal performance at $k=1$, implying that focused symptom selection maximizes discrimination while avoiding redundancy.

\section{Conclusion}

In this work, we present \tool{}, a framework that formalize multi-step clinical diagnosis as an active, knowledge-guided, and iterative refinement process. 
By integrating a medical KG for hypothesis grounding, an information gain–driven inquiry strategy for diagnostic uncertainty reduction, and a structured diagnostic record aligned with clinical standards, \tool{} enables systematic and efficient differential diagnosis in multi-turn dialogues. 
Unlike existing approaches that rely on static retrieval, heuristic questioning, or ungrounded LLM reasoning, our framework explicitly models the evolving clinical state and optimizes each diagnostic step toward maximal discriminative power. Experimental results demonstrate that \tool{} achieves both superior diagnostic accuracy and dialogue efficiency.

\section*{Limitations}

While \tool{} demonstrates promising performance in differential diagnosis, there are several limitations that warrant discussion. 
First, our patient simulation relies on LLM-generated case descriptions that may not fully capture the ambiguity of real patient narratives. 
Critically, our patient agent assumes cooperative and coherent symptom reporting, reflecting an idealized clinical interaction. 
In reality, patients often exhibit cognitive or linguistic biases: underreporting stigmatized symptoms, inaccurate recall, or anxiety-driven concerns rather than physiological reasoning.
In addition, our information gain computations assume conditional independence among symptoms given a disease and employ uniform likelihood over symptom–disease edges in KG. This assumption may lead to suboptimal question selection when diseases are distinguished primarily by complex symptom patterns.

\section*{Acknowledgements}

The application of AI in diagnosis support has raised ethical concerns that we carefully acknowledge.
\tool{} is proposed as a diagnostic reasoning assistant rather than a replacement for licensed doctors. All outputs must be validated by human doctors before any clinical action is taken.
Our evaluation data are derived from publicly available and anonymized datasets agent-MedQA, agent-CMB and NEJM Image Challenge. No real patient records are used, ensuring compliance with privacy standards.

\bibliography{references,references_manual}
\bibliographystyle{acl_natbib}
\clearpage
\appendix

\section{KG Implementation}
\label{kg}

To provide a clinically grounded foundation for our framework, we utilize PrimeKG (Precision Medicine Knowledge Graph) \cite{primekg2023}, which is a comprehensive resource that integrates over 20 high-quality primary sources, including Orphanet, Mayo Clinic, and DrugBank. 

PrimeKG comprises:

\begin{itemize}
	\item Nodes: Approximately 17,000 disease entities and 1,300 symptom entities.
	\item Edges: We prioritize two primary relationship types:
	\begin{itemize}
		\item Disease-Symptom: Indicating clinical manifestations associated with specific pathologies.
		\item Disease-Disease: Representing comorbidity links and hierarchical relationships (\eg ``is-a'' or ``part-of'' relations) that assist in differential grouping.
	\end{itemize}
\end{itemize}

\section{Detailed Description of Baselines}
\label{appendix1}
We provide a comprehensive overview of the 12 baseline methods used in our experiments:
\begin{itemize}
	\item \textbf{Dialog-Based Methods.} \textbf{AgentClinic} \cite{agentclinic2024samuel}: Simulates clinician–patient dialogues for diagnosis.
	CoT (Chain-of-Thought): Appends ``Let’s think step by step'' to encourage explicit reasoning.
	Huatuo \cite{huatuo2023wang} and  \cite{meditron2023chen}: Representative specialized medical LLMs.
	\item \textbf{KG-Based Methods.} 
	MCTS-BT: Uses Monte Carlo Tree Search with backtracking for hypothesis refinement.
	MCTS-MV: Extends MCTS by ranking symptom queries based on contextual informativeness.
	UoT (Unit of Thought) \cite{uot2024hu}: Constructs symptom-centric ``units'' around confirmed positive symptoms and prioritizes structural importance. 
	\item \textbf{KG-Based Methods.} 
	MEDIQ \cite{mediq2024li}: A diagnostic agent implementing multiple diagnostic strategies through sequential dialogues.
	CoD \cite{cod2025chen}: Selects questions by maximizing information entropy over candidate diseases.
	DDO \cite{ddo2025jia}: a multi-agent framework that dynamically chooses symptoms using diverse strategies.
	MEDDxAgent \cite{meddxagent2025rose}: Adapts questioning strategy based on diagnostic uncertainty.
	\item \textbf{SFT-Based Methods.} SFT / SFT-GT: Fine-tuning on Qwen3-8B using generated dialogues by AgentClinic~\cite{agentclinic2024samuel}, with or without ground-truth labels respectively.
\end{itemize}

\section{Dataset and Case Generation}

\subsection{Prompt for OSCE Case Generation}

We use the following prompt to generate standardized Objective Structured Clinical Examination (OSCE) scenarios for evaluation.
The prompt instructs the LLM to produce a structured JSON containing patient demographics, symptom history, physical findings, test results, and the ground-truth diagnosis while providing only the clinical objective to the Doctor Agent.

\begin{figure}[htbp]
	\caption{Prompt Template for OSCE Case Generation.}
	\begin{promptlisting}
		Please generate a sample Objective Structured Clinical Examination (OSCE) for the patient actor and the doctor, including what the correct diagnosis should be as a structured JSON. 
		
		Only provide the doctor with the objective and provide "test results" as a separate category. Provide these for a primary care doctor exam.
		
		Generate an OSCE for the following case study. Please read the answer category for the correct diagnosis. Here is an example of correcting the OSCE format (*@\placeholder{example}@*). Please create a new one here:
	\end{promptlisting}
	\label{prompt1}
\end{figure}

An example output is shown in Figure \ref{prompt2}.

\section{Implementation Details}

\subsection{Base Model and Inference Configuration}

- \textbf{Base models}: Qwen3-8B, Meta-Llama-3.1-8B-Instruct, and 

- \textbf{Inference temperature}: 0.05

- \textbf{Max tokens}: 2048

\subsection{Knowledge Graph Statistics}

We use PrimeKG~\cite{primekg2023}, which contains:

\quad- 17,080 disease nodes

\quad- 3,357 symptom nodes

\quad- 1,361 disease–disease relationships

\quad- 11,072 disease–symptom relationships

\subsection{LoRA Fine-Tuning Hyperparameters}
We fine-tune the base LLM using Low-Rank Adaptation (LoRA)~\cite{lora2021hu} with the following configuration:

\quad- \textbf{Learning rate}: 2e-4 

\quad- \textbf{Batch size}:

\qquad- \texttt{per\_device\_train\_batch\_size = 2}

\qquad- \texttt{gradient\_accumulation\_steps = 4}

\qquad- \texttt{Effective batch size = 2 × 4 = 8}

\quad- \textbf{LoRA rank (\texttt{r})}: 8

\quad- \textbf{Target modules}: \texttt{["q\_proj", "k\_proj", "v\_proj", "o\_proj"]}

\quad- \textbf{LoRA alpha}: 32

\quad- \textbf{Dropout}: 0.1

\quad- \textbf{Training epochs}: 3

\quad- \textbf{Warmup steps}: 1,000

\section{Prompt for Entity Extraction and Evaluation}

\paragraph{Symptom Entity Extraction.}
For symptom extraction from patient utterances, we employ the following prompt:

\begin{figure}[htbp]
	\caption{Prompt Template for Symptom Entity Extraction.}
	\begin{promptlisting}
		You are a helpful assistant with expertise in medical symptom identification.
		Please identify and extract all disease and symptom entities from the following sentence.
		Each entity must be no longer than 5 characters.
		
		Rules:
		1. Include symptoms that are affirmed (positive) in the "positive" list
		2. Include symptoms that are explicitly denied (negative) in the "negative" list
		3. Pay special attention to negation words like "no", "not", "don't", "haven't", "can still", which typically indicate negative symptoms
		
		Return the result in valid JSON format as shown below (without any markdown formatting and explanation):
		{
			"positive": ["Symptom 1", "Symptom 2", ...],
			"negative": ["Symptom 3", "Symptom 4", ...]
		}
		
		Sentence: (*@\placeholder{sentence}@*).
	\end{promptlisting}
	\label{prompt2}
\end{figure}

\paragraph{Diagnostic Accuracy Judgment.}
To evaluate whether the Doctor Agent’s final diagnosis matches the ground truth, we use the judgment prompt:

\begin{figure}[H]
	\caption{Prompt Template for Diagnosis Accuracy Check.}
	\begin{promptlisting}
		You are responsible for determining if the correct diagnosis and the doctor's diagnosis are the same disease. Please respond only with Yes or No. Nothing else.
		Here is the correct diagnosis: (*@\placeholder{ground truth diagnosis}@*)
		Here was the doctor diagnosis: (*@\placeholder{doctor diagnosis}@*)
		Are these the same?
	\end{promptlisting}
	\label{prompt3}
\end{figure}

\section{Prompt for Diagnostic Record Initialization and Update}

To ensure consistency of diagnostic record throughout the diagnosis process, we use a prompt that guides the LLM to perform evidence-based updates:

\begin{figure}[htbp]
	\caption{Prompt Template for Diagnostic Record Initialization and Update.}
	\begin{promptlisting}
		You are an experienced medical scribe.  
		Your task is to read the patient's latest utterance and incrementally update the structured summary below.
		Rules:
		1. Add or revise only facts confirmed in the CURRENT utterance.
		2. Preserve all existing information that is not contradicted.
		3. Use the exact JSON schema that was provided (do not create new keys).
		4. Return only the updated JSON object, with no extra commentary (without any markdown formatting).
		
		Schema: (*@\placeholder{schema}@*)
		Current structured summary: (*@\placeholder{current diagnostic record}@*)
		Latest patient utterance: (*@\placeholder{schema}@*)
		
		
	\end{promptlisting}
	\label{prompt4}
\end{figure}

\section{Prompt for Agent Initialization and Diagnosis}
\label{appendix6}

To simulate realistic clinical interactions, we implement a multi-agent diagnostic framework comprising three specialized agents: \Doctor, \Patient, and \Measurement with each guided by prompts to enforce specific behaviors and constraints. 

The \Doctor adopts a constrained prompt specifying question limits $T_{max}$ and tracks the count of asked questions $t_{current}$. 

\begin{figure}[H]
	\caption{Prompt Template for \Doctor Initialization.}
	\begin{promptlisting}
		You are a doctor named Dr. Agent who only responds in the form of dialogue. You are inspecting a patient whom you will ask questions in order to understand their disease.
		You are allowed to ask (*@\placeholder{T\_max}@*) questions total before you must make a decision, and have asked (*@\placeholder{t\_current}@*) questions so far. 
	\end{promptlisting}
	\label{prompt5}
\end{figure}

Additionally, during the \Doctor’s differential diagnosis, we employ the following prompt template, which integrates patient demographics, recent dialogue history, structured clinical findings, and relevant medical knowledge extracted from the knowledge graph.

\begin{figure}[H]
	\caption{Prompt Template for \Doctor Differential Diagnosis.}
	\begin{promptlisting}
		Age: (*@\placeholder{age}@*)
		Gender: (*@\placeholder{gender}@*)
		Chief Complaint: (*@\placeholder{chief complaint}@*)
		
		Recent dialogue history: (*@\placeholder{3 recent dialogue turns}@*)
		
		Medical record
		Chief Complaint: (*@\placeholder{chief complaint}@*)
		Symptoms: (*@\placeholder{symptoms}@*)
		Recent medical examinations: (*@\placeholder{recent medical examinations}@*)
		
		Knowledge Graph Context
		Relevant medical knowledge represented as triples:(*@\placeholder{simplified subgraph triples}@*)
		
	\end{promptlisting}
	\label{prompt6}
\end{figure}

The \Patient prevents patients from revealing diagnostic results directly, forcing the \Doctor to make diagnoses through symptom reasoning.

\begin{figure}[H]
	\caption{Prompt Template for \Patient Initialization.}
	\begin{promptlisting}
		You're a patient in a clinic. The doctor will ask questions or order exams to figure out your illness. 
		Below is all of your information: (*@\colorbox{blue!10}{\textcolor{blue}{\texttt{\{patient profile\}}}}@*)
		Never name your disease and only describe symptoms naturally: how they feel, when they flare, or how they affect you.
	\end{promptlisting}
	\label{prompt7}
\end{figure}

The \Measurement adopts a standardized result output format \texttt{(RESULTS: [results here])}, ensuring parseability of medical examination results.

\begin{figure}[H]
	\caption{Prompt Template for \Measurement Initialization.}
	\begin{promptlisting}
		You are a measurement reader who responds with medical test results. Please respond in the format "RESULTS: [results here]".
		Below is all of the information you have: (*@\colorbox{blue!10}{\textcolor{blue}{\texttt{\{medical examinations\}}}}@*). If the requested results are not in your data, then you can respond with NORMAL READINGS.
	\end{promptlisting}
	\label{prompt8}
\end{figure}

\begin{figure*}[htbp]
	\caption{An example of OSCE Case.}
	\begin{promptlisting}
		\quad
		{
			"OSCE Examination": {
				"Objective for Doctor": "Assess and diagnose the patient presenting with acute abdominal pain.",
				"Patient Actor": {
					"Demographics": "30-year-old female",
					"History": "The patient complains of sudden onset of sharp, right lower quadrant abdominal pain since last night. The pain has progressively worsened over the last 12 hours. She mentions that she felt nauseous this morning but has not vomited. No recent changes in bowel habits or urinary symptoms have been reported.",
					"Symptoms": {
						"Primary Symptom": "Sharp, right lower quadrant abdominal pain",
						"Secondary Symptoms": ["Nausea", "No vomiting", "No change in bowel habits", "No urinary symptoms"]
					},
					"Past Medical History": "No significant past medical history. No previous surgeries.",
					"Social History": "Non-smoker, occasional alcohol use. Works as a software developer.",
					"Review of Systems": "Denies fever, vomiting, diarrhea, dysuria, or flank pain."
				},
				"Physical Examination Findings": {
					"Vital Signs": {
						"Temperature": "37.2 (*@$^\circ$@*)C (99 (*@$^\circ$@*)F)",
						"Blood Pressure": "120/75 mmHg",
						"Heart Rate": "78 bpm",
						"Respiratory Rate": "16 breaths/min"
					},
					"Abdominal Examination": {
						"Inspection": "No distension or visible masses.",
						"Auscultation": "Normal bowel sounds.",
						"Percussion": "Tympanic throughout, no shifting dullness.",
						"Palpation": "Tenderness in the right lower quadrant. No guarding or rebound tenderness. Rovsing's sign positive, suggesting peritoneal irritation."
					}
				},
				"Test Results": {
					"Complete Blood Count": {
						"WBC": "12,000 /(*@$\mu$@*)L (elevated)",
						"Hemoglobin": "13.5 g/dL",
						"Platelets": "250,000 /(*@$\mu$@*)L"
					},
					"Urinalysis": {
						"Appearance": "Clear",
						"WBC": "2-5 /HPF",
						"RBC": "0-2 /HPF",
						"Nitrites": "Negative",
						"Leukocyte Esterase": "Negative"
					},
					"Imaging": {
						"Ultrasound Abdomen": {
							"Findings": "Enlarged appendix with wall thickening and fluid collection. No evidence of ovarian cyst or ectopic pregnancy."
						}
					}
				},
				"Correct Diagnosis": "Acute Appendicitis"
			}
		}
		
		
	\end{promptlisting}
	\label{prompt9}
\end{figure*}

\end{document}